# Adversarial Training for EM Classification Networks

Tom F. Grimes[1], Eric D. Church[1], William K. Pitts[1], Lynn S. Wood[1], Eva Brayfindley[1], Luke Erikson[1], Mark Greaves[1]

*Abstract*—We present a novel variant of Domain Adversarial Networks with impactful improvements to the loss functions, training paradigm, and hyperparameter optimization. New loss functions are defined for both forks of the DANN network, the label predictor and domain classifier, in order to facilitate more rapid gradient descent, provide more seamless integration into modern neural networking frameworks, and allow previously unavailable inferences into network behavior. Using these loss functions, it is possible to extend the concept of 'domain' to include arbitrary user defined labels applicable to subsets of the training data, the test data, or both. As such, the network can be operated in either 'On the Fly' mode where features provided by the feature extractor indicative of differences between 'domain' labels in the training data are removed or in 'Test Collection Informed' mode where features indicative of difference between 'domain' labels in the combined training and test data are removed (without needing to know or provide test activity labels to the network). This work also draws heavily from previous works on Robust Training which draws training examples from a L∞ ball around the training data in order to remove fragile features induced by random fluctuations in the data. On these networks we explore the process of hyperparameter optimization for both the domain adversarial and robust hyperparameters. Finally, this network is applied to the construction of a binary classifier used to identify the presence of EM signal emitted by a turbopump. For this example, the effect of the robust and domain adversarial training is to remove features indicative of the difference in background between instances of operation of the device – providing highly discriminative features on which to construct the classifier. This technique is proving useful across a wide range of scientific applications.

*Index Terms*— STFT-CNN, DANN, TCI-DANN, Robust Training, LIME, Interpretability

## I. INTRODUCTION

CONSTRUCTING neural classifiers on data using established neural network techniques (e.g. Convolutional Neural Networks (CNNs) [1]) has a known issue of reliance on feature representations that are correlated with background [2]. This is perhaps even more common for networks modeling scientific data with extremely poor signal-to-noise quality. During attempts to improve network performance on these real-world datasets derived from active laboratory environments, two promising techniques have emerged for alleviating this issue: Robust Training [3] and Domain Adversarial Neural Networks (DANNs) [4]. Both approaches work towards generating features that are uncorrelated with background and thus more capable of correct classification through a range of likely test environments.

The task highlighted in this paper for the demonstration of these techniques is that of the construction of a binary classifier to identify the on/off state of a turbopump via voltage data. Specifically, the device used for this experiment is a Pfeiffer HiCube 400 vacuum pump (shown in Figure 1) chosen as a readily available, easily movable, and simply operable exemplar of lab equipment at Pacific Northwest National Laboratory (PNNL). The voltage data was gathered from a wall-circuit by a Saleae Logic Pro 16. This data acquisition (DAQ) system operated on 3 channels at 1.56 MS/s/channel. The channels corresponded to Hot-Ground, Hot-Neutral, and Ground-Neutral.

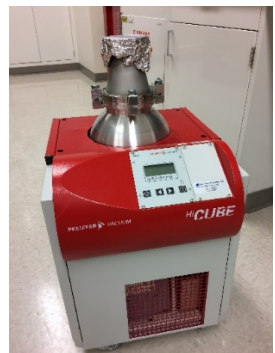

Fig. 1: HiCube 400 Classic

Several datasets were constructed with the DAQ and HiCube at varying separation distances (both physically and electrically). As the distances increased, the signal-to-noise ratio of the signal emitted by the turbopump decreased. The data was taken in an active laboratory with a large number of workers engaged in operating both office and lab equipment. This activity induced different 'background' conditions which classifiers needed to be robust against. From closest to farthest, these locations were:

- On different circuits connected in the same panelboard
- On different circuits feeding into panelboards, transformers, and a common switchboard
- On different circuits feeding into panelboards, transformers and switchboards connected to outside power in adjacent buildings

The circuit diagrams for these scenarios are summarized in Figure 2.

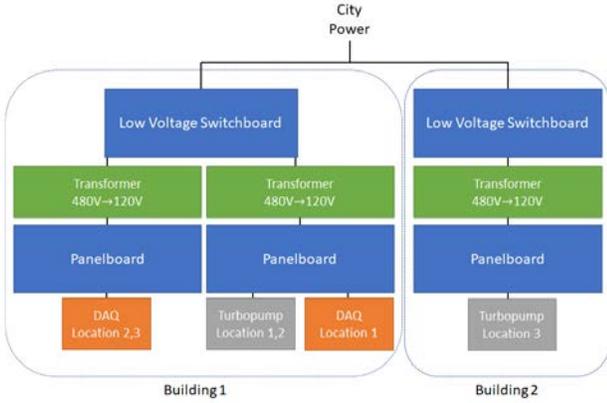

Fig. 2: Experiment circuit diagram summary

The voltage data collected by the DAQ was preprocessed into a spectrogram by applying a Short time Fourier Transform (STFT). Each of the 3 channels of data were individually transformed and formed 3 planes of a single input. The duration of the spectrogram was set to 0.1s of data (1.56E5 samples at 1.56 MHz).

Spectrograms were generated by randomly selecting a class, a collection corresponding to that class, and a starting point in the data. Training data was selected from the first ~80% of data files. After a buffer to prevent overlap, validation data was selected from the end ~20% of the files. Test data was generated from a smaller set of files set aside during training.

In order to classify the STFTs, a standard feature extractor was used – ResNet18 [5]. The SciPy [6] STFT algorithm was configured to directly generates output in the 224x224 shape of the input required input into the feature extractor.

Explanations for the model output were constructed using LIME [7]. LIME (Local Interpretable Model-agnostic Explanations) works as a technique to identify the pieces of an input that were influential to the classification of that input by any black-box model. Linear weights are fit to the model's prediction on near permutations of the input being explained and used to identify the most important input features to classification. By aggregating over many LIME outputs, it is possible to identify frequency bands in the STFT representation that are the most important to classification.

Further details on the data preparation and feature extraction section of the network as well as the explanation techniques can be found in [8].

## II. ADVERSARIAL TECHNIQUES

### A. Adversarial Nomenclature

In "Adversarial examples are not bugs, they are features" [9] Ilyas et. al. observe that the reason that image classifiers are vulnerable to misclassification induced by imperceptible changes to the input is not a shortcoming of the network, but rather it is the result of imperceptible features being present and useful for prediction in the data. In fact, classification based on these features is transferable between models of very different architectures that are trained over the same dataset.

In order to further the discussion of this concept, the nomenclature and definitions from that paper have been adopted:

Features will be rigorously defined as a mapping from the input dimension onto the set of real numbers:

$$F = \{f : \mathbb{R}^n \to \mathbb{R}^1\}$$

A feature $f$ will be defined as ρ-useful if the feature is correlated with the true label in expectation $\mathbb{E}$ for input x and true label y over domain D:

$$\mathbb{E}_{(x,y)\sim D}[y \cdot f(x)] \geq \rho$$

A feature $f$ will be defined as γ-useful if the feature is correlated with the true label as well as the true label of all other inputs that can be reached in the neighborhood of the example with attack budget, Δ. For this paper, the neighborhood reachable with Δ will be limited by the magnitude of the $L_\infty$ distance from the input. Thus, γ-useful features are defined by:

$$\mathbb{E}_{(x,y)\sim D}\left[\inf_{\delta \in \Delta}(y \cdot f(x+\delta))\right] \geq \gamma$$

Classification is performed using a weighted linear combination of features. For a binary classifier:

$$C(x) = sgn\left(b + \sum_{f \in F} w_f \cdot f(x)\right)$$

Classifiers are willing to exploit any ρ-useful features in order to maximize accuracy. This includes the non-γ-useful features that are not human perceivable and are human-indistinguishable from noise when displayed.

Observed in the data taken for this paper, ρ-useful but not γ-useful features taken from the training domain are unlikely to be correlated in expectation with the labels for the validation and test data. As such, it is desirable to find means of training that removes these features and incorporates only γ-useful features.

### B. Robust Training

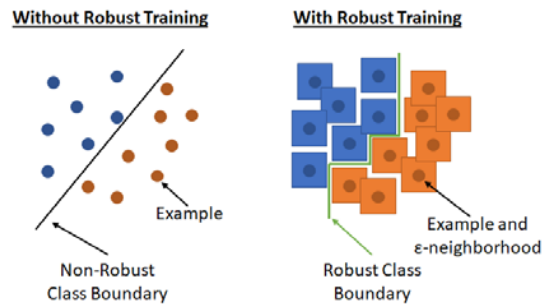

Fig. 3: Illustration of Robust Training Methods

Robust training is a means for removing features that are ρ-useful but not γ-useful from the model. During training, instead of training on examples that are taken directly from the dataset of interest, the examples are modified before being used for training [3]. For images with the value of the input normalized to [0,1], a new uniform distribution on the interval [-ε, ε] is added to every pixel across all 3 planes and the resultant is clipped to fit in [0,1]. New values for the distribution are generated for every pixel of every example every epoch when the image is presented to the training algorithm.

As such, the network is in effect selecting inputs from a neighborhood around the data point within a fixed $L_\infty$ distance, $\varepsilon$. Only features correlated in expectation with the true label for all the inputs in this neighborhood will remain after training. This precisely matches our definition of $\gamma$-usefulness. Thus, robust training is an effective means for attempting to produce $\gamma$-useful features.

Unfortunately, this process is not done for free. Removing $\rho$-useful features is extremely likely to generate an attendant loss in training accuracy [10]. The features being removed are meaningful and predictive on the training set despite the fact that they may not transfer to the test set.

Figure 3 shows the effect of robust training on the classification boundary. As examples are trained from the region surrounding the examples, the boundary between classes shifts such that all examples within the adversarial region around the example are classified the same way.

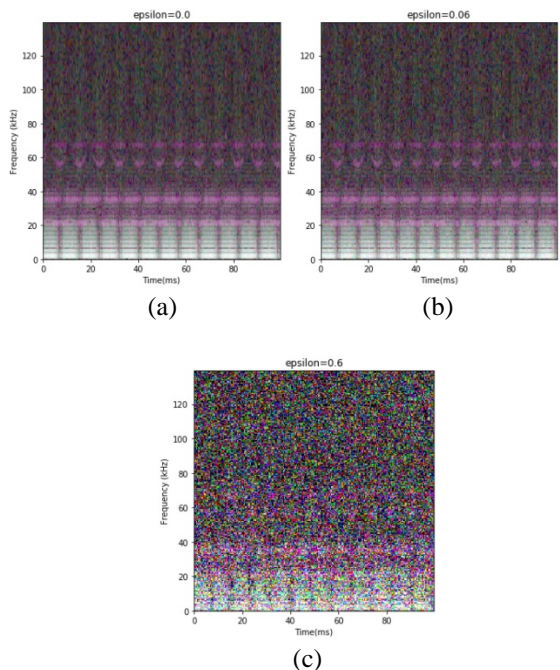

(a)

(b)

(c)

Figure 4a,b,c: Input data sampled from the $L_\infty$ neighborhood around a data point with $\varepsilon = 0$, 0.06, 0.6

Figure 4 shows how robust training impacts the input that is given to the network. Figure 4a shows an unmodified input corresponding to epsilon=0. Figure 4b and 4c show how the input is modified by adding a uniform noise distribution with epsilon=0.06 and 0.6 respectively. The difference between Figures 4a and 4b is extraordinarily hard to see by eye. It will be shown during subsequent sections (see the discussion of the Robust Training Hyperparameter) that this amount of disruption is sufficient for disturbing some $\rho$-useful features induced by backgrounds and statistical fluctuations but is insufficient for disturbing a useful number of $\gamma$-useful features corresponding to signal from the device of interest.

*C. Domain Adversarial Networks*

In addition to wanting to remove $\rho$-useful but not $\gamma$-useful features from the model, there is an additional set of features that are providing accuracy in the training set that do not translate into features that are useful in the test set. These are the $\gamma$-useful features that are associated with the background. These features are removed using Domain Adversarial Neural Networks (DANN). For this work we will use 'domain' to indicate any variable for which it is desired that the network should generate uncorrelated features (e.g. a variable for collection # which is used to indicate categorical information about the background environment)

The key insight that led to the usage of the DANN is that the features that are useful for class identification across all domains would not be useful for identifying the difference between domains. However, without intervention there are likely to be features that are positively correlated with a class label that are also positively correlated with some subset of domain labels. The ideal network is one that would generate features that are strongly correlated with the class label but have no correlation with the domain label. Thus, the architecture of the DANN is designed to create such features.

Figure 5 shows a schematic of the network design. Typical image classification (e.g. using ResNet on ImageNet) would consist of using the feature extractor and attaching the label predictor to the top feature layer to perform the classification task. To use a DANN, the domain classifier fork is added to the network and likewise attached to the top feature layer of the feature extractor. Between the feature extractor and the domain classifier is inserted a gradient reversal layer.

The total loss is calculated as the sum of the loss from the label predictor and domain classifier. Both use categorical cross-entropy (or a variant thereof). Updates to the label predictor and domain classifier are done as normal in order to minimize the loss of their respective forks. Backpropagation flowing through the label predictor to the feature extractor also flows as normal. However, the gradient reversal layer causes updates made to the feature extractor based on the domain loss to be made in the opposite direction.

Although moving in the opposite of the direction that leads to the most accurate discrimination between the domain of the input isn't the same as moving in the direction that leads to the least discriminating capability, in practice this process eventually leads to the generation of features at the top level of the feature extractor that have no information that is useful for domain classification. As such, the domain classifier is no more or less accurate than random guessing.

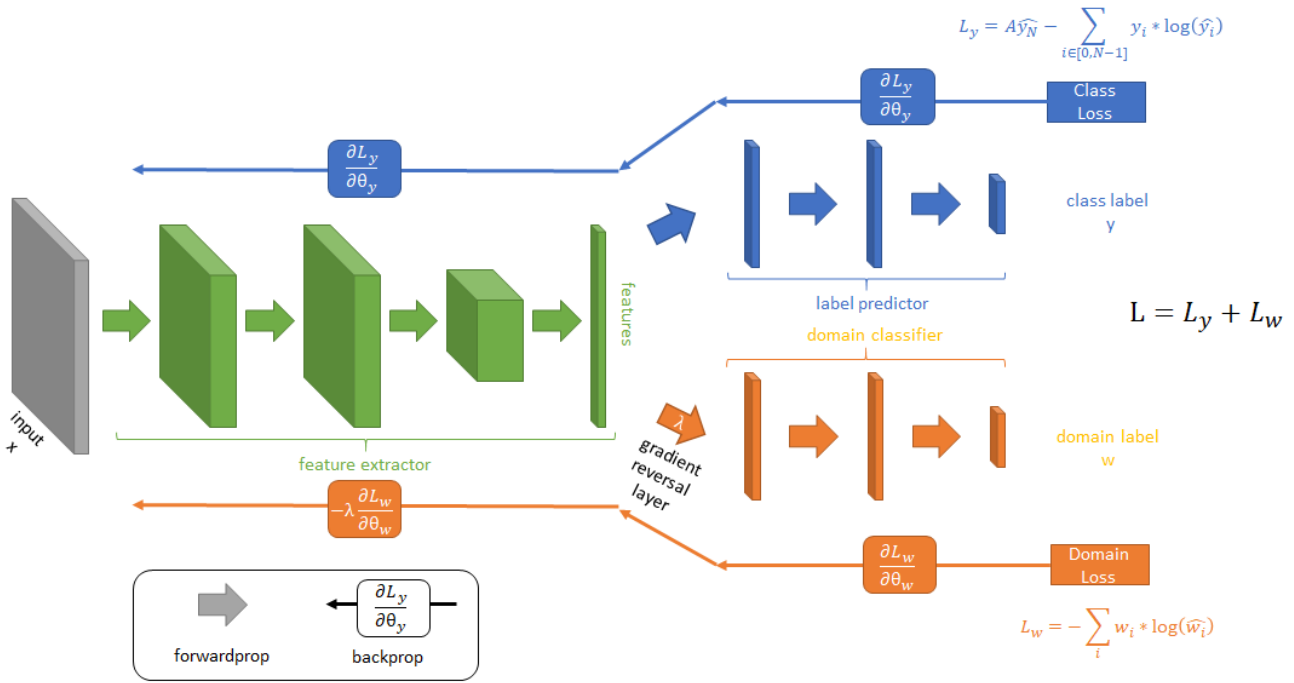

Figure 5: Test Collection Informed Domain Adversarial Network Schematic

The gradient reversal layer is controlled by a hyperparameter $\lambda$. Setting $\lambda > 0$ allows the gradient reversal process to proceed as described - minimizing the available domain information. Setting $\lambda < 0$ causes updates to the feature extractor to be in the direction that maximize domain label classification (as if there was no reversal layer). Setting $\lambda = 0$ causes the feature extractor to not be updated by the loss function from the domain classifier and the network behaves as if it were not attached (the domain classifier is still updated). Increases in $|\lambda|$ cause the importance of domain updates to increase in relative proportion to the importance of class label updates.

Training with this DANN architecture can proceed in two distinct modes: On-the fly (OTF) and Test Collection Informed (TCI). See Figure 6a and 6b.

On-the-fly training is relatively straightforward. Both the label predictor and domain classifier are trained using simple categorical cross-entropy loss on the training data. It has the advantage in that it can be precomputed and used for inference as test data is collected. However, it relies on the assumption that the backgrounds evidenced in the training set are sufficiently representative of the full set of background conditions such that feature representations uncorrelated with the backgrounds in the training data will also be uncorrelated with the backgrounds in the test data.

For Test Collection Informed training, the test data is added to the training data during training in order to generate features that are also uncorrelated with domain labels in the test data. On its face using test data during training seems to violate fundamental principles of supervised learning. However, because the network is only provided with the input data and the domain label and NOT the class label, there is no leakage of information about the output of interest into the training of the model.

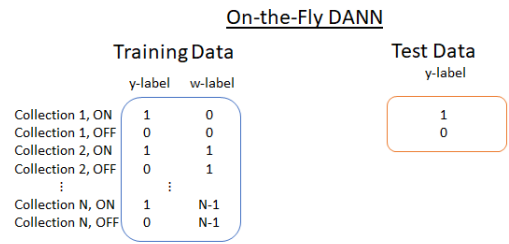

Figure 6a: On-The-Fly Domain Adversarial Network

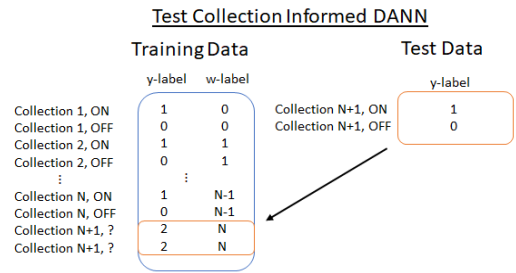

Figure 6b: Test Collection Informed Domain Adversarial Network

The mechanics for training a TCI-DANN require generating a new class label that signifies the classification of the test data is 'unknown' during training. The domain label is still used to identify the state of the variable that is desired to be uncorrelated with the features being used for classification.

In addition to re-labeling the data, the loss function must also be updated. The domain loss remains categorical cross-entropy. Gradient reversed updates on the domain loss from both training and test inputs will result in features at the top layer of the feature extractor that are uncorrelated with any of the domain labels that are present in either the train or test sets.

The loss from the label classifier, however, needs to be modified in order to prevent the network from updating based on the classification label of inputs from the test set and to prevent the network from predicting that inputs belong the unknown class. The modified loss function is the following:

$$L_y = A\widehat{y_N} - \sum_{i \in [0,N-1]} y_i * \log(\widehat{y_i})$$

Where $\sum_{i \in [0,N-1]} y_i * \log(\widehat{y_i})$ represents standard categorical cross-entropy across all known classes, $\widehat{y_N}$ represents the prediction weight put on the unknown class, and $A$ represents a large constant penalty term (e.g. A=500) used to prevent the network from allocating prediction mass to the unknown class. Using this loss function, prediction of the unknown class is quickly extinguished by the penalizing term. Once that happens, the loss reverts to standard categorical cross-entropy for all inputs with known labels. When presented with unknown input, the network will not update between known class labels because there is no gradient in the loss between classes.

Once training is complete, the same test data used to contribute domain information during training can be fed into the network to perform class prediction. Due to the modified label predictor loss function, all inputs will predict one of the known classes. The domain fork can be detached once the weights are no longer being updated to save on calculations during inference.

The DANN network as described here contains many meaningful departures from Yaroslav et al [4] that have substantial impact on the efficiency and usefulness of the architecture.

First, the concept of domain has been broadened from referencing either the 'training' or 'test' set to referencing states of any variable for which it is desired that the features used by the network have no correlation. On-the-fly DANN formulations become possible due to this redefinition because the domain variable can take several values in the training set. For variables with many possible domain states, the DANN process becomes more likely to identify and remove features indicative of background due to an increased number of distinct sets to compare across. Using this type of domain variable, the DANN becomes relevant to a broad class of scientific problems where it is very common to have data obtained from collections in several discrete background conditions.

Second, the relabeling scheme and modified class loss makes it possible to train both forks of the network simultaneously. This procedure is easier to implement using modern deep learning frameworks and produces better gradient updates for faster, more efficient training by using batches with a mix of inputs from many domains and computing updates from both forks simultaneously.

Finally, by changing the overall loss to be a direct sum of the loss of the branches (rather than multiplying domain loss by the adversarial hyperparameter), there is meaningful diagnostic information that is generated by the network (this is described in the discussion of the domain adversarial hyperparameter)

## III. Hyperparameter Optimization

### A. Robust Training Hyperparameter, ε

For many data sets, the solution at ε=0 (not using robust training) achieves extremely high training accuracy, but poor validation and test accuracy. As previously discussed, this is because the network is using ρ-useful but not not γ-useful features that are present in the training data but are not correlated in expectation with the true label in the validation and test sets In order to combat this issue, robust training was employed.

As ε is increased from 0, the network begins to be prevented from using a subset of ρ-useful features that are not γ-useful (alternately defined as features that are γ-useful but only for small attack budgets, Δ). As a result, the training accuracy decreases because the network is no longer able to use these features to improve the accuracy during training. However, the model adapts by making predictions using new and remaining γ-useful features. These features are not quite as effective at predicting the training set (else, these features would be more prominently used by the ε=0 model), but they are more useful at predicting the validation and test sets and lead to an increase in accuracy for those predictions.

Eventually, increases in ε will reach the point that real γ-useful features that truly pervade all the training/validation/test sets will begin to be obscured from the training. The point at which a unit increase in ε has no impact on the validation accuracy (because the impacts of removing true features and false features are equalized) is the optimal point for operation, ε*. Beyond this point, increases in ε do more to damage to true features than they do good by removing false features. ε* is depicted in Figure 5. For most datasets in this paper ε*≈0.06 as obtained by grid search.

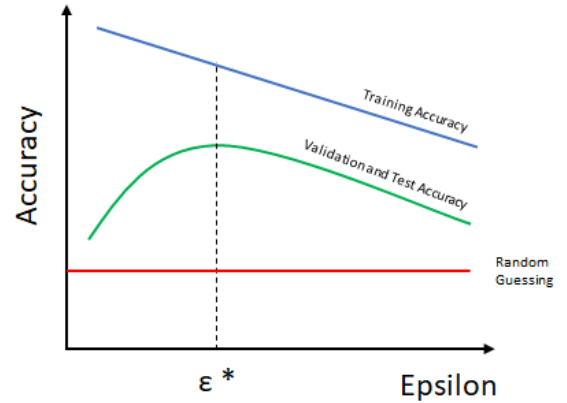

Figure 7: Optimization of the Robust Training Hyperparameter

### B. Domain Adversarial Hyperparameter, λ

As previously discussed, when training with λ = 0 and using the overall loss function presented in this work the DANN backpropagation will still make changes to the domain classifier, but those changes will not propagate backwards to the feature extractor. As such, it is possible to find the extent to which a CNN (DANN with λ = 0) uses features that are domain indicative and thus likely to be influenced by changes in background.

What we find is that as the CNN network trains it arrives at a set of features at the top layer of the feature extractor that contains information that allows the domain classifier to make very accurate domain predictions as shown in Figure 8. This is the feature set that the network would arrive at if the DANN fork were not attached. While these CNN features allow the label predictor to make accurate label predictions using the training data or data taken in very similar backgrounds, the strong predictive power of the features on the domain label makes it likely that the classifier would fail if the background were to change significantly.

For DANN networks with $\lambda = 1$, however, the features selected by the feature extractor are constrained to features that are not correlated with the domain. Thus, the domain classifier is unable to do better than random guessing when attempting to predict the domain. The features selected in this manner are therefore more likely to be robust to changes in background. Domain accuracy prediction curves (like Figure 8) for $\lambda = 0.5$ and 500 appear similar to the $\lambda = 1$ curve in that both values seem sufficient to substantially remove domain information from the features identified by the feature extractor. However, the impact on the accuracy of the label predictor fork of the network will likely differ.

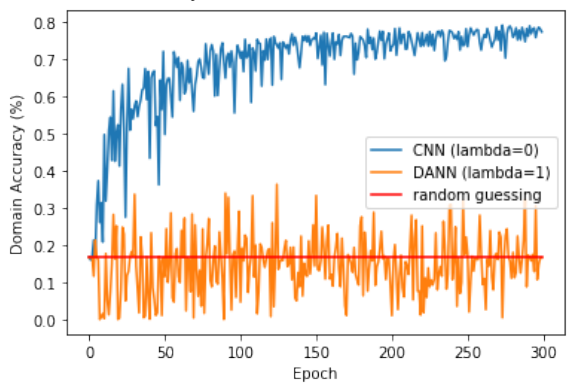

Figure 2: Domain prediction accuracy for CNN (DANN with $\lambda = 0$) and DANN (with $\lambda = 1$) trained with cross-transformer data (section IV-B) representing 6 domain conditions

Just as the $\varepsilon$ hyperparameter needs to be optimized for label prediction accuracy, so too does the $\lambda$ hyperparameter. First, the value of $\varepsilon^*$ is found and fixed for the network. With $\lambda=0$, the training optimizes for the training set at the expense of the validation and test sets by using all the background features that are available without penalty. As $\lambda$ increases, the magnitude of the updates to the feature extractor from the domain loss increases. As such, features that are correlated with one or more domain labels are preferentially removed from the feature extractor.

Thus, the training accuracy decreases because the network is no longer able to use these features during training. However, the model adapts by using new and remaining features that are less correlated with domain labels and thus more useful for prediction on the validation and test sets and the attendant accuracy increases.

Eventually, $\lambda$ will increase to the point that updates to the feature extractor from the domain loss will reduce the classification performance not only for the training set but also for the validation and test sets by aggressively removing features that exhibit extremely narrow correlation in expectation with a domain label (despite actually being a feature correlated in expectation with the true classification label across all domains). The point of transition where a unit increase in $\lambda$ causes no change in the validation accuracy is identified as $\lambda^*$ and depicted in Figure 9. Values of $\lambda$ were tested between .5 and 500. Values of $\lambda^*$ used in this paper will be either 1 or 0.5.

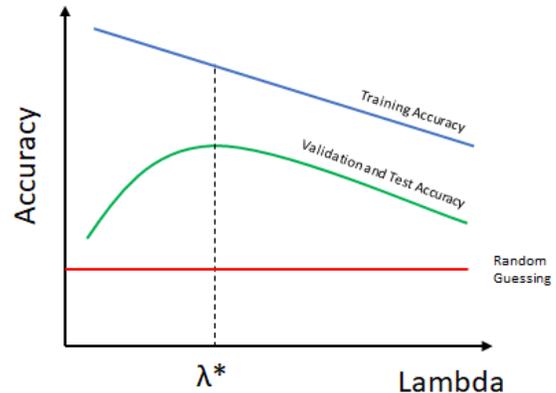

Figure 3: Optimization of the Domain Adversarial Hyperparameter $\lambda$

## IV. RESULTS

As briefly described in the introduction and depicted in Figure 2, the results in this paper detail 3 experiments: Cross-Panelboard (location 1), Cross-Transformer (location 2), and Cross-Building (location 3)

Because the validation data is more proximate in time to the training data (it was taken from a different section of the same collection) than the test data (it was taken from a different collection), the validation set shares more background features with the training set than the test set does. This explains the consistent lag that shows up between validation and test accuracy.

It is also worth noting that the results in the tables don't reflect the accuracy of the full system, but rather the accuracy for the classification of an individual spectrogram. Spectrograms built closely in time have been shown to be correlated in their prediction but are not perfectly correlated even for spectrograms starting at the very next sample of the DAQ. By aggregating classification results over many spectrogram classifications, the system result becomes much more accurate than the individual spectrogram classification. Given 1.56E6 spectrograms that can be generated per second, even individual accuracy rates slightly over 50% are likely to rapidly converge to correct identifications of equipment state.

For all the models constructed in the following section, the LIME explainability code and aggregation technique described in [7] was used to identify the sections of input important to the model decision and thereby identify the frequencies used by the model to identify the state of the turbopump.

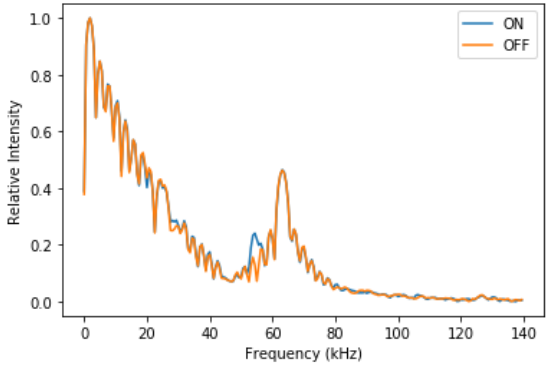

Figure 10: Sample FFTs for On and Off states of Cross-Panelboard data

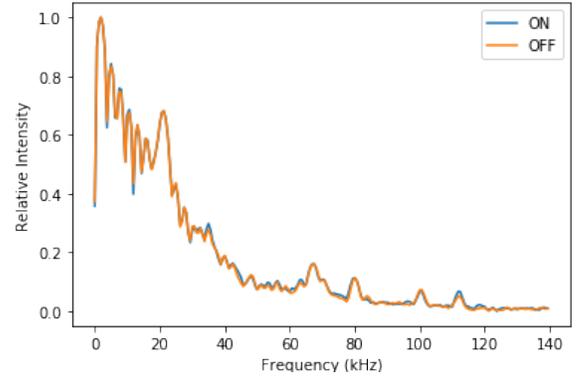

Figure 12: Sample FFTs for On and Off states of Cross-Transformer data

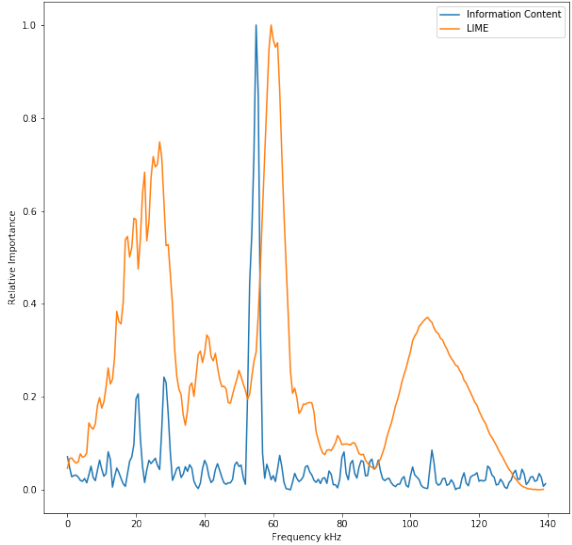

Figure 11: Cross-Panelboard On the fly DANN model frequency usage predicted by LIME vs frequencies emitted by the turbopump (absolute difference between curves in Figure 10)

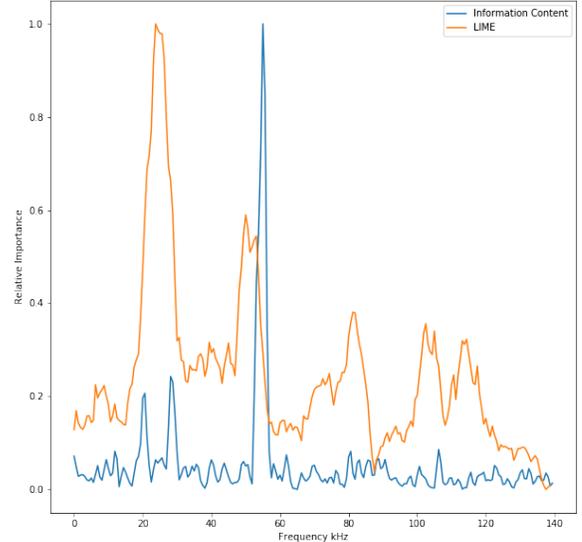

Figure 13: Cross-Panelboard On the fly DANN model frequency usage predicted by LIME vs frequencies emitted by the turbopump (absolute difference between curves in Figure 10)

TABLE I: Cross-Panelboard Results

| Lambda | Epsilon | TCI-DANN? | Train Acc | Val Acc | Test Acc |
|--------|---------|-----------|-----------|---------|----------|
| **0** | 0 | 0 | 100 | 100 | 100 |
| **1** | 0 | 0 | 100 | 100 | 100 |

TABLE II: Cross-Transformer Results

| Lambda | Epsilon | TCI-DANN? | Train Acc | Val Acc | Test Acc |
|--------|---------|-----------|-----------|---------|----------|
| **0** | 0 | 0 | 100 | 87.19 | 76.20 |
| **1** | 0 | 0 | 99.9 | 89.60 | 74.40 |
| **1** | 0.06 | 0 | 95.77 | 91.77 | 83.00 |

## A. Cross-Panelboard

The first set of data was obtained with the DAQ operated on a different circuit attached to the same GE Spectra SBO panelboard as the turbopump. Six repetitions were performed of the process whereby 30 seconds of data was taken with the turbopump on followed as soon as possible by 30 seconds of data with the turbopump off. Five repetitions were used for training with the first 90% being used for training and the last 10% held back for validation. One pair chosen at random was held back for test data.

Figure 10 shows sample FFTs corresponding to samples taken from 'on' and 'off' data in this configuration. For this data it is possible to separate the turbopump states by eye. Particularly there is a large peak between 50 and 60 kHz that is recognizable.

A simple CNN with no adversarial techniques was generally able to distinguish between states with 100% accuracy. This is unsurprising for inputs that are easily distinguishable by eye. On-The-Fly Domain Adversarial techniques were also applied, and the results were fairly similar both in prediction accuracy and in frequency usage identified by LIME.

Figure 12 compares the absolute value of the difference of the FFTs from on and off states in Figure 10 to the frequencies identified by LIME as being used by the On-the-fly DANN network for discrimination.

This difference between the FFTs in the on and off states with the turbopump in the closest location to the DAQ represents the purest available approximation for frequencies of interest emitted by the turbopump. This will therefore be used for all the models to compare to frequencies LIME identifies as important to the model. For the Cross-Panelboard model, as depicted in Figure 11, LIME indicates that the network generally seems to be using frequencies that a human would identify as useful discriminating between these states. Accuracy results for the CNN and the On-The-Fly DANN networks on this data are displayed in Table 1.

### B. Cross-Transformer

The second set of data was obtained for the DAQ and Turbopump positioned physically and electrically as far away as they could be in the same building. This meant each was connected to a circuit leading into a GE Spectra SBO panelboard which was in turn connected to a common GE Low Voltage Switchboard AXD-10 with an interposed 480 to 120V transformer.

For this dataset, again 6 sets of on and off data were obtained, but the captures were extended to 300s. Sample FFTs for on and off are shown in Figure 12. The sets are no longer distinguishable by eye. It is also notable that there was a significant change in the background of the building between collections due to both a change in DAQ location and changes in background equipment operation.

The results from this set began to show meaningful improvement from the application of adversarial techniques. Particularly of note, nonzero values of ε and λ chosen empirically to be near the optimal of the curve for validation accuracy resulted in improved accuracy on the validation and test sets while simultaneously reducing the accuracy on the training set. This is just as would be predicted from Figures 7 and 9. Results are shown in Table 2.

The areas that LIME identified as most important for classification by this network were very similar to the areas identified as important to the Cross-Panelboard dataset (see Figure 13).

### C. Cross-Building

The third set of data was obtained with the DAQ in the same location as the Cross-Transformer experiment and turbopump positioned in a nearby building. As before, the DAQ and turbopump were connected to outside power via a panelboard, transformer, and switchboard. Details of the circuit connections outside the buildings are currently unknown.

For this dataset, 12 sets of on and off data were obtained – again with 300s captures. As with the Cross-Transformer data the FFTs are no longer distinguishable by eye (Figure 14).

Despite the DAQ being located in the same spot on the same circuit, the background changed due to different operation patterns of equipment in the building. Importantly, the FFT in Figure 14 appears to indicate that there was a device located very close to the DAQ that induced signals in the same frequency bands as the most dominant frequency of the turbopump (~60 kHz). The interfering unknown device was in the on state for all the data that was collected.

Several small changes were made for the models that addressed this dataset. The spectrogram frequency axis was adjusted to have a range of 0 to 80 kHz rather than 0 to 140 kHz to help provide more focus on the interesting frequencies 20 to 60 kHz. This change had a minimal difference on performance, but theoretically could have allowed the network to become sensitive to features in the frequency direction that were previously being averaged over with the larger frequency window. Additionally, the procedure for optimization of λ* used different node points and therefore recovered a value of 0.5 instead of 1. Finally, in order to improve on the statistics each of the network types were run 20 times with different training data sampled from the same experiments. Accuracy values in the Table III represent the mean of the 20 networks. Because 20 networks were run for each row in Table III it was possible to assemble a 'hard' ensemble by combining the results of all of the networks and assigning the result corresponding to the most frequent response. This result appears in the Ensemble Acc column of Table 3.

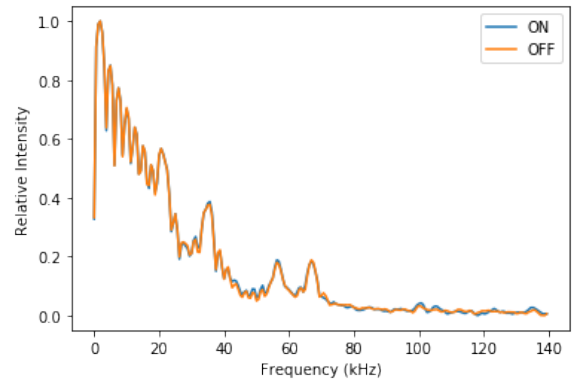

Figure 14: Sample FFTs for On and Off states of Cross-Building data

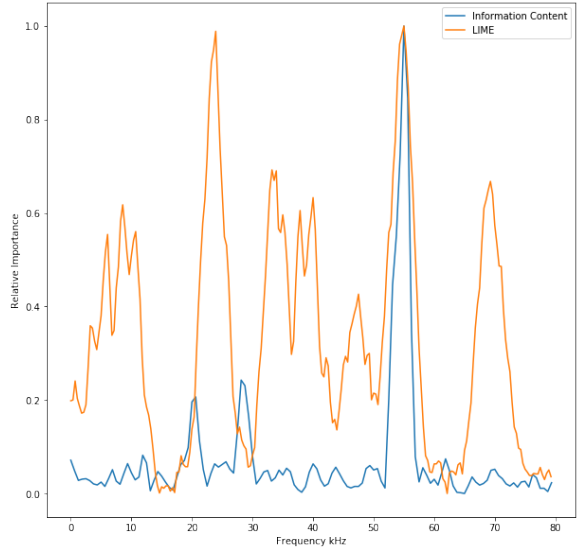

Figure 15: Highest accuracy Cross-Building TCI-DANN model frequency usage predicted by LIME vs frequencies emitted by the turbopump (absolute difference between curves in Figure 10)

TABLE III: Average Cross-Building Results

| Lambda | Epsilon | TCI-DANN? | Train Acc | Val Acc | Test Acc | Ensemble Acc |
|---|---|---|---|---|---|---|
| 0 | 0 | 0 | 92.19 | 69.62 | 61.34 | 63.10 |
| 0.5 | 0.06 | 0 | 84.87 | 66.31 | 64.56 | 65.45 |
| 0.5 | 0.06 | 1 | 84.76 | 69.86 | 67.97 | 72.80 |

Again, as expected, the CNN significantly outperformed the DANN variants in training accuracy. Also, as before, the validation accuracy evaluated on data proximate in time and highly correlated in background features to the training data performed very similarly between the CNN and DANN variants. However, the DANN networks proved superior in identifying background robust features and therefore outperformed the CNN on the test set where the background was less correlated due to large difference in the time of the collection. The TCI-DANN networks perform particularly well due to their ability to remove the effects of background not only in the training set, but also in the test set. Figure 16 shows the full distribution of test accuracies evaluated at the epoch corresponding to the highest validation accuracy for each of the models.

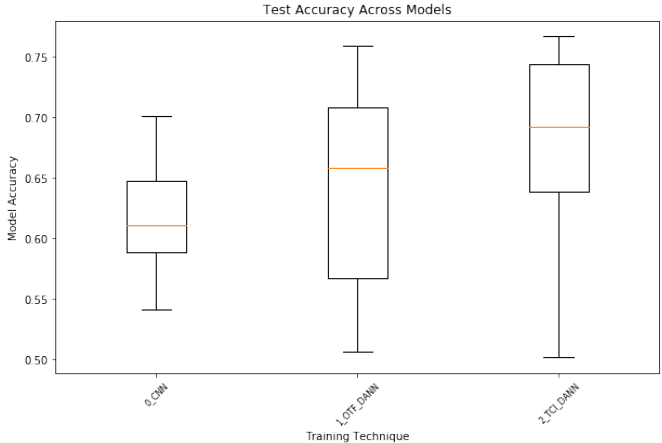

Figure 16: Distribution of test accuracy across models for Cross-Building

While the difference in the performance between the CNN and the TCI-DANN may not seem overwhelming, the ~10 percentage point increase in accuracy from the CNN ensemble to the TCI-DANN ensemble represents a condition where more than ~25% of the time the TCI-DANN ensemble will correctly classify a spectrogram when the CNN ensemble would classify incorrectly. This increase in accuracy makes the full system relying on many spectrograms converge much faster and more accurately

Despite the extreme reduction in the signal to noise ratio, the frequencies identified by LIME in the highest accuracy TCI-DANN model remain indicative of frequencies found in the models for the previous experiments (Figure 15). This provides additional confidence that the model has properly been able to accommodate this much more difficult task.

## V. CONCLUSION

In this paper we have explored the usefulness of robust training and domain adversarial neural networks (DANNs) to allow the rejection of features that are unlikely to translate to successful characterization of data outside the training set. In identifying these features, we find that thinking in $\rho$-useful and $\gamma$-useful features is helpful.

We find that our default approach of using what we term On-the-fly-DANN models, in which the feature extractor is conditioned to reject features indicative of domain in the training data is sufficient for providing resilience to backgrounds similar to those that exist in the training data. This methodology proved sufficient for identifying turbopump signal across multiple transformers.

Finally, and most powerfully, we find that the Test Collection Informed DANN (TCI-DANN) variant holds the extremely useful property of being robust to changes in background associated with the test data – the data that neural models are most interested in predicting correctly. This is accomplished by causing updates to the feature extractor from the domain of the test data without providing or training on the class label of the data. Using this methodology, it was possible to detect turbopump activity through multiple transformers and in a different building than the data acquisition system.

Finally, neural explainability techniques in the form of LIME were able to show the responsiveness of the model feature usage to the frequencies most likely to be useful for distinguishing activity. This provides additional evidence for the strength of the models generated through the chosen techniques.

A summary of all the results is provided in Table 4.

TABLE IV: Full Result Summary

| Scenario | Lambda | Epsilon | TCI-DANN? | Train Acc | Val Acc | Test Acc |
|---|---|---|---|---|---|---|
| Cross-Panelboard | 0 | 0 | 0 | 100 | 100 | 100 |
| Cross-Panelboard | 1 | 0 | 0 | 100 | 100 | 100 |
| Cross-Transformer | 0 | 0 | 0 | 100 | 87.19 | 76.20 |
| Cross-Transformer | 1 | 0 | 0 | 99.9 | 89.60 | 74.40 |
| Cross-Transformer | 1 | 0.06 | 0 | 95.77 | 91.77 | 83.00 |
| Cross-Building | 0 | 0 | 0 | 92.19 | 69.62 | 61.34 |
| Cross-Building | 0.5 | 0.06 | 0 | 84.87 | 66.31 | 64.56 |
| Cross-Building | 0.5 | 0.06 | 1 | 84.76 | 69.86 | 67.97 |

## VI. FUTURE WORK

The techniques described in this paper are showing tremendous promise across a wide spectrum of neural modeling tasks for scientific data. The scenario where data is obtained in many different collections with different sets of background features that must be removed is extremely common in scientific applications and the ability to overcome that issue opens many doors to new lines of scientific inquiry.

Additionally, these techniques may have significant ramifications for AI fairness. Having shown that the concept of domain can be extended from referencing train and test to encompassing any arbitrary label, it is possible to train a network in such a manner that the features used for decision making by the network are completely uncorrelated with whatever variable it is desired for the network not to take into account. This could have powerful implications for fields where guaranteeing networks do not use certain attributes of input data in support of their inferences is a continued subject of intense research.

## ACKNOWLEDGEMENT

This work was funded by the U.S. Department of Energy National Nuclear Security Administration's Office of Defense Nuclear Nonproliferation Research and Development (NA-22).

We wish to acknowledge our Pacific Northwest National Laboratory colleagues for their important discussions: Aaron Tuor, Alex Hagen, and Aaron Luttman.

## VII. BIOGRAPHIES

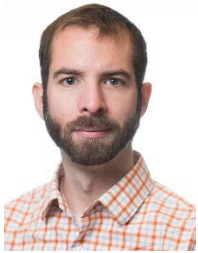

**Tom Grimes** received his PhD in Nuclear Engineering and MBA from Purdue University in 2015.

He is a Physicist at PNNL. Prior to working at PNNL, he was a Postdoctoral Fellow and an NSF Graduate Fellow at Purdue University.

Dr. Grimes' research interests focus on deep learning with an emphasis on explainability. Applications of interest include electromagnetics, particle spectroscopy, and metastable fluids.

**Eric Church** received his PhD from the University of Washington, Seattle, WA, in Experimental Particle Physics in 1996.

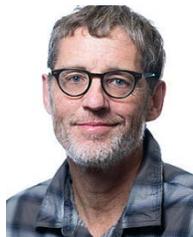

He is a Physicist at PNNL. Prior to his arrival at PNNL, he was a Research Associate Scientist with Yale University and a post-doctoral researcher with UC-Riverside.

Dr. Church leads various internal lab investments and DOE projects including research into neutrino oscillations and the majorana nature of neutrinos. Current experimental neutrino collaborations on which he works include MicroBooNE, DUNE and NEXT.

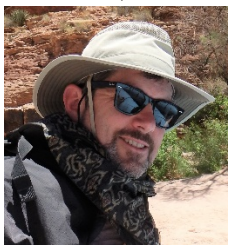

**William Karl Pitts** received his PhD in Nuclear Physics from Indiana University in 1987. He is a Physicist at PNNL. Prior to joining PNNL in 2000 he was an Associate Professor from the University of Louisville and a Research Scientist at the University of Wisconsin. His current interests include signal processing, EM/RF studies, and material processing.

**Lynn Wood** received his PhD from the University of California, Davis in Nuclear Physics in 1998, after which he was a post-doctoral researcher at Iowa State University before spending 11 years in the embedded systems industry, where he managed ASIC processor design development. He is currently a Physicist at PNNL.

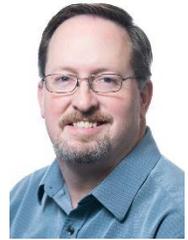

Dr. Wood leads PNNL's involvement in the Belle II high-energy physics experiment based in Japan, as well as several projects sponsored by National Nuclear Security Administration.

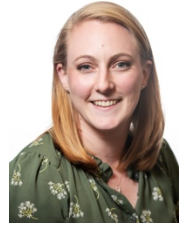

**Eva Brayfindley** received her PhD in Applied Mathematics from North Carolina State University in 2019.

She is a data scientist at PNNL with research interests in machine learning with an emphasis on both explainability and uncertainty quantification. Applications of interest include mass spectrometry, metabolomics and cheminformatics.

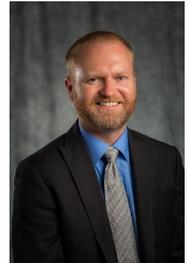

**Luke Erikson** is a Scientist in the National Security Directorate at Pacific Northwest National Laboratory with broad experience in computer science and nuclear physics.

He earned his Ph.D. in Applied Nuclear Astrophysics from the Colorado School of Mines.

**Mark Greaves** is currently Technical Director for Analytics in the National Security Directorate at PNNL.

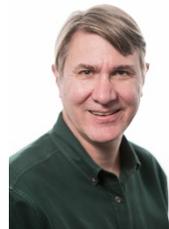

Prior to coming to PNNL, Mark was Director of Knowledge Systems at Vulcan Inc., Director of DARPA's Joint Logistics Technology Office, and Program Manager in DARPA's Information Exploitation Office. Mark began his career at Boeing, where he worked on advanced programs in software agent technology.

He has published two books and over 40 papers, holds two patents, chaired the FIPA technical committee on agent communications languages, and was awarded the Office of the Secretary of Defense Medal for Exceptional Public Service for his contributions to US national security while serving at DARPA. Mark holds a PhD from Stanford University.